\documentclass[a4paper,twoside]{article}
\usepackage[T1]{fontenc}
\usepackage[utf8]{inputenc}
\usepackage[english]{babel}
\usepackage{times}
\usepackage{import}
\usepackage{graphicx}
\usepackage{subfigure}
\usepackage{booktabs}
\usepackage{amsmath,amsfonts,bm}
\usepackage{appendix}
\usepackage{hyperref}
\usepackage{algorithm}
\usepackage{algorithmic}

\import{JFPDA/}{jfpda2020.sty}

\DeclareMathOperator*{\E}{\mathbb{E}}
\renewcommand{\eqref}[1]{Equation (\ref{#1})}
\newcommand{\secref}[1]{Section \ref{#1}}
\newcommand{\appref}[1]{Appendix \ref{#1}}
\newcommand{\conference}{icml}

\begin{document}
\title{\Large\bf End-to-end learning of reusable skills through intrinsic motivation}

\author{\begin{tabular}[t]{c@{\extracolsep{6em}}c@{\extracolsep{6em}}c}
A. Aubret${}^1$ & L. Matignon${}^2$ & S. Hassas${}^1$\\
\end{tabular}
{} \\
 \\
${}^1$        Univ Lyon, Université Lyon 1, CNRS, LIRIS F-69622, Villeurbanne, France
{} \\
 \\
arthur.aubret@univ-lyon1.fr\\
}

\maketitle

\subsection*{Abstract}
{\em
Humans beings do not think about the muscles they contract but rather take high level decisions such as walking or running. Taking inspiration from developmental learning, we present a novel reinforcement learning architecture which hierarchically learns and represents self-generated skills in an end-to-end way. With this architecture, an agent focuses only on task-rewarded skills while keeping the learning process of skills bottom-up. This bottom-up approach allows to learn skills independently from extrinsic reward and transferable across tasks. To do that, we combine a previously defined mutual information objective with a novel curriculum learning algorithm, creating an unlimited and explorable tree of skills. We test our agent on a simple gridword environment to understand and visualize how the agent distinguishes between its skills. Then we show that our approach can scale on more difficult  MuJoCo environments in which our agent is able to build a representation of skills which facilitates transfer learning.
}
\subsection*{Keywords}
Intrinsic motivation, curriculum learning, developmental learning, reinforcement learning.

\section{Introduction}

In reinforcement learning (RL), an agent learns by trial-and-error to maximize the expected rewards obtained from actions performed in its environment \cite{sutton1998reinforcement}. 
However, many RL agents usually strive to achieve one goal using only low-level actions. In contrast, as humans being, when we want to go to work, we do not think about every muscle we contract in order to move; we just take abstract decisions such as \textit{Go to work}. Low-level behaviors such as how to walk are already learned and we do not need to think about them. Learning to walk is a classical example of babies developmental learning, which refers to the ability of an agent to spontaneously explore its environment and acquire new skills \cite{barto2013intrinsic}. Babies do not try to get walking behaviors all at once, but rather first learn to move their legs, to crawl, to stand up, and then, eventually, to walk. They are \textbf{intrinsically motivated} since they act for the inherent satisfaction of learning new skills \cite{ryan2000intrinsic} rather than for an \textbf{extrinsic reward} assigned by the environment. 

Several works are interested in learning abstract actions, also named \textbf{skills} or \textbf{options} \cite{sutton1999between}, in the framework of deep reinforcement learning (DRL) \cite{aubret2019survey}.
Skills can be learned with extrinsic rewards \cite{bacon2017option}, which  facilitates the credit assignment \cite{sutton1999between}. 
In contrast, if one learns skills with intrinsic motivation, the learning process becomes bottom-up \cite{machado2017laplacian}, i.e. the agent learns skills before getting extrinsic rewards. When learning is bottom-up, the agent commits to a time-extended skill and avoids the usual wanderlust due to the lack, or the sparsity, of extrinsic rewards. Therefore, it can significantly improve exploration \cite{machado2016learning,nachum2019does}. In addition, these skills can be used for different tasks, emphasizing their potential for transfer learning \cite{taylor2009transfer}. These properties make intrinsic motivation attractive in a \textit{continual learning} framework, which is the ability of the agent to acquire, retain and reuse its knowledge over a lifetime \cite{thrun1996learning}.

Several works recently proposed to intrinsically learn such skills using a diversity heuristic     \cite{eysenbach2018diversity,achiam2018variational}, such that different states are covered by the learned skills. 
Yet several issues remain: 1- the agent is often limited in the number of learned skills or requires \textit{curriculum learning} \cite{achiam2018variational}; 2- most skills target uninteresting parts of the environment relatively to some tasks; thereby it requires prior knowledge about which features 
to diversify \cite{eysenbach2018diversity}; 3- the agent suffers from catastrophic forgetting when it tries to learn a task while learning skills \cite{eysenbach2018diversity}; 4- discrete time-extended skills used in a hierarchical setting  are often sub-optimal for a task. With diversity heuristic, skills are indeed not expressive enough to efficiently target a goal \cite{eysenbach2018diversity,achiam2018variational}.

In this paper, we propose to address these four issues so as to \textbf{improve the approaches for continually learning increasingly difficult skills with diversity heuristics}. We introduce ELSIM (End-to-ended Learning of reusable Skills through Intrinsic Motivation), a method for learning representations of skills in a bottom-up way. The agent autonomously builds a tree of abstract skills where each skill is a refinement of its parent. First of all, skills are learned independently from the tasks but along with tasks; it guarantees they can be easily transferred to other tasks and may help the agent to explore its environment. We use the optimization function defined in \cite{eysenbach2018diversity} which guarantees that states targeted by a skill are close to each other. 
Secondly, the agent selects a skill to refine with extrinsic or intrinsic rewards, and learns new sub-skills; it ensures that the agent learns specific skills useful for tasks through an intelligent \textit{curriculum}, among millions of possible skills. 

Our approach contrasts with existing approaches which either bias skills towards a task \cite{bacon2017option}, reducing the possibilities for transfer learning, or learn skills during pretraining \cite{eysenbach2018diversity}. We believe our paradigm, by removing the requirement of a \textit{developmental period} \cite{metzen2013incremental} (which is just an unsupervised pretraining), makes naturally compatible developmental learning and \textit{lifelong learning}.
Therefore, we emphasize three properties of our ELSIM method. 1-\textbf{Learning is bottom-up}: the agent does not require an expert supervision to expand the set of skills. It can use its skills to solve different sequentially presented tasks or to explore its environment. 2- \textbf{Learning is end-to-end}: the agent never stops training and keeps expanding its tree of skills. It gradually self-improves and avoids catastrophic forgetting. 3- \textbf{Learning is focused}: the agent only learns skills useful for its high-level extrinsic/intrinsic objectives when provided. 


Our contributions are the following: we introduce a new curriculum algorithm based on an adaptation of diversity-based skill learning methods. Our objective is not to be competitive when the agent learns one specific goal, but \textbf{to learn useful and reusable skills along with sequentially presented goals in an end-to-end fashion}. 
We show experimentally that ELSIM achieves \textbf{good asymptotic performance} on several single-task benchmarks, \textbf{improves exploration} over standard DRL algorithms and manages to easily \textbf{reuse its skills}. Thus, this is a step towards \textit{lifelong learning} agents.

This paper is organized as follows. First, we introduce the concepts used in ELSIM, especially 
diversity-based intrinsic motivation (\secref{sec:background}). In \secref{sec:method}, the core of our method is presented. 
Then, we explain and visualize how ELSIM works on simple gridworlds and compare its performances with state-of-the-art DRL algorithms on single and sequentially presented tasks learning  
(\secref{sec:expe}). 
In \secref{sec:related}, we detail how ELSIM relates to existing works. Finally, in \secref{sec:conclusion}, we take a step back and discuss ELSIM.

\section{Background} \label{sec:background}

\subsection{Reinforcement learning}
 A Markov decision process (MDP) \cite{puterman2014markov} is defined by a set of possible states $S$; a set of possible actions $A$; a transition function $P : S \times A \times S \rightarrow \mathbb{P}(s'|s,a)$ with $a\in A$ and $s,s' \in S$; a reward function $R : S \times A \times S \rightarrow \mathbb{R}$; the initial distribution of states $\rho_0 : S \rightarrow [0;1]$. A stochastic policy $\pi$ maps states to probabilities over actions 
 in order to maximize the discounted cumulative reward defined by $\varsigma_t=\left[\sum_{t=0}^{\infty} \gamma^t r_t\right]$ where $\gamma \in[0,1]$ is the discount factor. In order to find the action maximizing $\varsigma$ in a state $s$, it is common to maximize the expected discounted gain following a policy $\pi$ from a state-action tuple defined by:

\ifthenelse{\equal{\conference}{ecml}}{

\begin{equation}
Q_{\pi}(s,a) = \E_{a{_t}\sim\pi(s{_t}),
s_{t+1}\sim P(s_{t+1}|s_t,a_t)}
\left[\sum_{t=0}^{\infty} \gamma{^t} R(s{_t},a{_t},s_{t+1})\right] \label{eq:espeQ}
\end{equation}
}{

\begin{equation}
Q_{\pi}(s,a) = \E_{\substack{a{_t}\sim\pi(s{_t})\\
s_{t+1}\sim P(s_{t+1}|s_t,a_t)}}
\left[\sum_{t=0}^{\infty} \gamma{^t} R(s{_t},a{_t},s_{t+1})\right] \label{eq:espeQ}
\end{equation}
}

where $s_0=s$, $a_0=a$. To compute this value, it is possible to use the Bellman Equation \cite{sutton1998reinforcement}.

\subsection{Obtaining diverse skills through mutual information objective}\label{sec:objmi}

One way to learn, without extrinsic rewards, a set of different skills along with their intra-skill policies is to use an objective based on mutual information (MI). In \cite{eysenbach2018diversity}, 
learned skills should be as \textbf{diverse} as possible (different skills should visit different states) and \textbf{distinguishable} (it should be possible to infer the skill from the states visited by the intra-skill policy). It follows that the learning process is 4-step with two learning parts \cite{eysenbach2018diversity}: 1- the agent samples one skill from an uniform distribution; 2- the agent executes the skill by following the corresponding intra-skill policy; 3- a discriminator learns to categorize the resulting states to the assigned skill; 4- at the same time, these approximations reward intra-skill policies (cf. \eqref{eq:int_reward}).

The global objective can be formalized as maximizing the MI between the set of skills $G$ and states $S'$ visited by intra-skill policies, defined by \cite{gregor2016variational}:
\ifthenelse{\equal{\conference}{ecml}}{

\begin{align}
    I(G;S') &= \mathbb{H}(G) - \mathbb{H}(G|S') =\mathbb{E}_{g\sim p(g),s'\sim p(s'|\pi_{\theta}^g,s)} [\log p(g|s')-\log p(g)] \label{eq:mi}
\end{align}
}{
\begin{align}
    I(G;S') &= \mathbb{H}(G) - \mathbb{H}(G|S') \\ &=\mathbb{E}_{\substack{g\sim p(g)\\s'\sim p(s'|\pi_{\theta}^g,s)}} [\log p(g|s')-\log p(g)] \label{eq:mi}
\end{align}
}

where $\pi^g_{\theta}$ is the intra-skill policy of $g\in G$ and is parameterized by $\theta$; 
$p(g)$ is the distribution of skills the agent samples on; and $p(g|s')$ is the probability to infer $g$ knowing the next state $s'$ and intra-skill policies. This MI quantifies the reduction in the uncertainty of $G$ due to the knowledge of $S'$. By maximizing it, states visited by an intra-skill policy have to be informative of the given skill. 


A bound on the MI can be used as an approximation to avoid the difficulty to compute $p(g|s')$\cite{barber2003algorithm,gregor2016variational} :
\ifthenelse{\equal{\conference}{ecml}}{
\begin{align}
    I(G,S') \geq \mathbb{E}_{g\sim p(g),s'\sim p(s'|\pi_{\theta}^g,s)}[\log q_{\omega}(g|s') -\log p(g)]\label{eq:bound}
\end{align}
}{
\begin{align}
    I(G,S') \geq \mathbb{E}_{\substack{g\sim p(g)\\s'\sim p(s'|\pi_{\theta}^g,s)}}[\log q_{\omega}(g|s') -\log p(g)]\label{eq:bound}
\end{align}
}
where $q_{\omega}(g|s')$ is the \textbf{discriminator} approximating $p(g|s')$. In our case, the discriminator is a neural network parameterized by $\omega$. $q_{\omega}$ minimizes the standard cross-entropy $- \mathbb{E}_{g\sim p(g|s')}\log q_{\omega}(g|s')$ where $s'\sim \pi^g_{\theta}$.

To discover skills, it is more efficient to set $p(g)$ to be uniform as it maximizes the entropy of $G$ \cite{eysenbach2018diversity}.  Using the uniform distribution, $\log p(g)$ is constant and can be removed from \eqref{eq:bound}. It follows that one can maximize \eqref{eq:bound} using an intrinsic reward to learn the intra-skill policy of a skill $g \in G$ \cite{eysenbach2018diversity}:
\begin{equation}
    r^g(s') = \log q_{\omega}(g|s').
    \label{eq:int_reward}
\end{equation}

Similarly to \cite{eysenbach2018diversity}, we use an additional entropy term to encourage the diversity of covered states. In practice, this bonus is maximized through the use of DRL algorithms: Soft Actor Critic (SAC) \cite{haarnoja2018soft} for continuous action space and Deep Q network (DQN) with Boltzmann exploration \cite{mnih2015human} for discrete one.

\section{Method}\label{sec:method}

In this section, we first give an overview of our method and then detail the building of the tree of skills, the learning of the skill policy, the selection of the skill to refine and how ELSIM integrates this in an end-to-end framework. We define a \textbf{skill} as a policy that target an area of the state space. This area is named the \textbf{goal} of the skill; a goal and a skill are deeply connected and used interchangeably in this paper.

\subsection{Overview: building a tree of skills}

To get both bottom-up skills and interesting skills relatively to some tasks, our agent has to choose the skills to improve thanks to the extrinsic rewards, but we want that our agent improves its skills without extrinsic rewards.
The agent starts by learning a discrete set of diverse and distinguishable skills using the method presented in \secref{sec:objmi}.
Once the agent clearly distinguishes these skills using the covered skill-conditioned states with its discriminator, it splits them into new sub-skills. For instance, for a creature provided with proprioceptive data, a \textit{moving forward} skill could be separated into \textit{running} and \textit{walking}. 
The agent only trains on sub-skills for which the parent skill is useful for the global task. Thus it incrementally refines the skills it needs to accomplish its current task. If the agent strives to sprint, it will select the skill that provides the greater speed. 
The agent repeats the splitting procedure until its skills either reach the maximum number of splits or become too deterministic to be refined.

The \textbf{hierarchy of skills} is maintained using a tree where each node refers to an abstract skill that has been split and each leaf is a skill being learned. We formalize the hierarchy using sequence of letters where a letter's value is assigned to each node:
\begin{itemize}
    \item The set of skills $G$ is the set of leaf nodes. A skill $g\in G$ is represented by a sequence of $k+1$ letters : $g=(l^0,l^1,...,l^k)$. When $g$ is split, a letter is added to the sequence of its new sub-skills. For instance, the skill $g=(l^0=0,l^1=1)$ can be split into two sub-skills $(l^0=0,l^1=1,l^2=0)$ and  $(l^0=0,l^1=1,l^2=1)$.
    \item The vocabulary $V$ refers to the values which can be assigned to a letter. For example, to refine a skill into 4 sub-skills, we should define $V=\{0,1,2,3\}$.
    \item The length $L(g)$ of a skill is the number of letters it contains. Note that the length of a skill is always larger than its parent's.
    \item $l^{:k}$ is the sequence of letters preceding $l^k$ (excluded).
\end{itemize}

We use two kind of policies: the first are the \textbf{intra-skill policies}. As a key element of each skill, they are described in \secref{sec:learn_skill}. The second type of policy is task-dependent and responsible to choose which skill to execute; we call it the \textbf{tree-policy} (see \secref{sec:tree_policy}).

\subsection{Learning intra-skill policies}\label{sec:learn_skill}

In this section, we detail how intra-skill policies are learned. We adapt the method presented in \secref{sec:objmi} to our hierarchical skills context. Two processes are simultaneously trained to obtain diverse skills: the intra-skill policies learn to maximize the intrinsic reward (cf. \eqref{eq:int_reward}), which requires to learn a discriminator $q_{\omega}(g|s')$. Given our hierarchic skills, we can formulate the probability inferred by the discriminator as a product of the probabilities of achieving each letter of $g$ knowing the sequence of preceding letters, by applying the chain rule: 

\ifthenelse{\equal{\conference}{ecml}}{

\begin{align}
    \centering
    r^g(s') &= 
    \log q_{\omega}(g|s')
        = \log q_{\omega}(l^0,l^1, \dots, l^k | s')\nonumber\\&= \log \prod_{i=0}^{k} q_{\omega}(l^{i} | s',l^{:i})
        = \sum_{i=0}^{k} \log q_{\omega}(l^{i} | s',l^{:i}). \label{eq:myintreward}
\end{align}

}{
\begin{align}
    \centering
    r^g(s') &= \log q_{\omega}(g|s')
        = \log q_{\omega}(l^0,l^1, \dots, l^k | s') \nonumber \\
        &= \sum_{i=0}^{k} \log q_{\omega}(l^{i} | s',l^{:i}). \label{eq:myintreward}
\end{align}
}

Gathering this value is difficult and requires an efficient discriminator $q_{\omega}$. As it will be explained in \secref{sec:catastrophic}, in practice, we use \textbf{one different discriminator for each node} of our tree: $\forall i,\  q_{\omega}(l^i|s',l^{:i})\equiv q_{\omega}^{:i}(l^i|s')$.

For instance, if $|V|=2$, one discriminator $q_{\omega}^{\emptyset}$ will be used to discriminate $(l^0=0)$ and $(l^0=1)$ but an other one, $q_{\omega}^{l^0=0}$ will discriminate $(l^0=0,l^1=0)$ and $(l^0=0,l^1=1)$.

It would be difficult for the discriminators to learn over all letters at once; the agent would gather states for several inter-level discriminators at the same time and a discriminator would not know which part of the gathered states it should focus on. This is due to the fact that discriminators and intra-skill policies simultaneously train. Furthermore, there are millions of possible combinations of letters when the maximum size of sequence is large. We do not want to learn them all. To address these issues, we introduce \textbf{a new curriculum learning algorithm that refines a skill only when it is distinguishable}. When discriminators successfully learn, they progressively extends the sequence of letters; in fact, we split a skill (add a letter) only when its discriminator has managed to discriminate the values of its letter. Let's define the following probability: 
\begin{equation}
p_{finish}^{:k}(l^k)= \mathbb{E}_{s_{final}\sim \pi^{:k+1}}\left[ q_{\omega}^{:k}(l^k|s_{final})\right].\label{eq:endlearn}
\end{equation}

where $s_{final}$ is the state reached by the intra-skill policy at the last timestep. We assume the discriminator $q_{\omega}^{:k}$ has finished to learn when: $\forall v \in V,\;  p_{finish}^{:k}(l^k=v) \geq \delta$ where $\delta \in [0,1]$ is an hyperparameter. Choosing a $\delta$ close to $1$ ensures that the skill is learned, but an intra-skill policy always explores, thereby it may never reach an average probability of exactly 1; we found empirically that 0.9 works well.

To approximate \eqref{eq:endlearn} for each letters' value $v$, we use an exponential moving average $p_{finish}^{:k}(l^k=v)=(1-\beta)p_{finish}^{:k}(l^k=v)+\beta q_{\omega}^{:k}(l^k=v|s_{final})$ where $s_{final} \sim \pi^{:k+1}$ and $\beta \in [0;1]$. Since we use buffers of interactions (see \secref{sec:catastrophic}), we entirely refill the buffer before the split.

Let us reconsider \eqref{eq:myintreward}. $\sum_{i=0}^{k-1} \log q_{\omega}(l^{i} | s',l^{:i})$ is the part of the reward assigned by the previously learned discriminators. It forces the skill to stay close to the states of its parent skills since this part of the reward is common to all the rewards of its parent skills. In contrast, $\log q_{\omega}(l^k|s',l^{:k})$ is the reward assigned by the discriminator that actively learns a new discrimination of the state space. Since the agent is constrained to stay inside the area of previous discriminators, the new discrimination is \textbf{uncorrelated} from previous parent discriminations. In practice, we increase the importance of previous discriminations with a hyper-parameter $\alpha \in \mathbb{R}$: 
\begin{equation}
r^g(s') = \log q_{\omega}(l^k|s',l^{:k})+\alpha  \sum_{i=0}^{k-1} \log q_{\omega}(l^{i} | s',l^{:i}).
\end{equation}

This hyper-parameter is important to prevent the agent to deviate from previously discriminated areas to learn more easily the new discrimination.

\subsection{Learning which skill to execute and train}\label{sec:tree_policy}

For each global objective, a stochastic policy, called \textit{tree-policy} and noted $\pi_T$ (with $T$ the tree of skills), is responsible to choose the skill to train by navigating inside the tree at the beginning of a task-episode. This choice is critical in our setting: while expanding its tree of skills, the agent cannot learn to discriminate every leaf skill at the same time since discriminators need states resulting from the intra-skill policies. We propose to \textbf{choose the skill to refine according to its benefit in getting an other reward} (extrinsic or intrinsic), thereby ELSIM executes and learns only interesting skills (relatively to an additional reward).


To learn the \textit{tree-policy}, we propose to model the tree of skills as an MDP solved with a Q-learning and Boltzmann exploration. The action space is the vocabulary $V$; the state space is the set of nodes, which include abstract and actual skills; the deterministic transition function is the next node selection; if the node is not a leaf, the reward function $R_T$ is $0$, else this is the discounted reward of the intra-skill policy executed in the environment divided by the maximal episode length. Each episode starts with the initial state as the root of the tree, the \textit{tree-policy} selects the next nodes using Q-values. Each episode ends when a leaf node has been chosen, i.e. a skill for which all its letters has been selected; the last node is always chosen uniformly (see \secref{sec:catastrophic}).

\begin{figure*}[t]
\centering
\includegraphics[width=9cm]{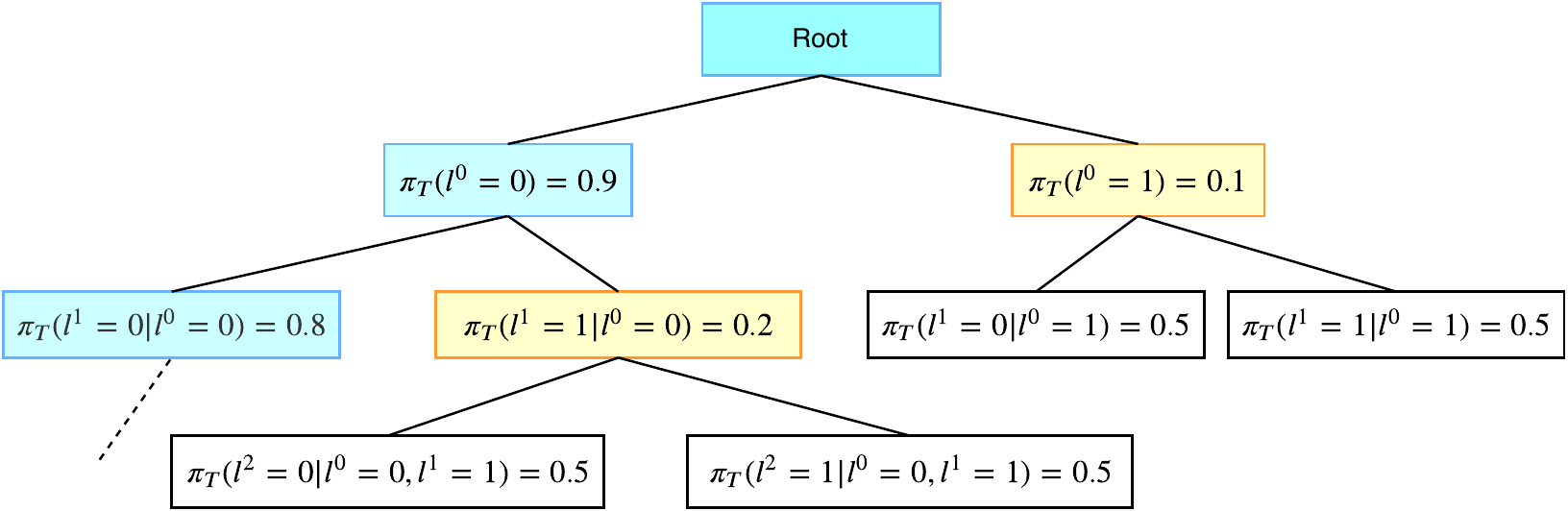}
\caption{Representation of a part of the tree of skills with $|V|=2$ and the value of \textit{tree-policy} in each node. 
White nodes are actual leaves of the tree; the discriminator is inactive. Yellow nodes represent nodes for which the discriminator can not differentiate its sub-skills; the \textit{tree-policy} samples uniformly. Nodes are blue when the discriminator can distinguish its sub-skills; the \textit{tree-policy} samples using Q-values.}\label{fig:tree}
\end{figure*}

Let us roll out an example using the \textit{tree-policy} displayed in Figure \ref{fig:tree}. The episode starts at the root of the tree; the \textit{tree-policy} samples the first letter, for example it selects $l^0=0$. Until it reaches a leaf-node, it samples new letters, e.g. $l^1=1$ and $l^2=0$. The \textit{tree-policy} has reached a leaf, thereby it will execute and learn the skill $(0,1,0)$. Then, the tuple $((0,1),(0))$ is rewarded with the scaled discounted reward of the task. This reward is propagated with the Bellman equation \cite{sutton1998reinforcement} to state-action tuples $((\emptyset),(0))$ and $((0),(1))$ to orientate the \textit{tree-policy} to $(0,1,0)$.



The MDP evolves during the learning process since new letters are progressively added. The Q-values of new skills are initialized with their parent Q-values. However, \eqref{eq:myintreward} ensures that adding letters at the leaf of the tree monotonically increases Q-values of their parent nodes. The intuition is that, when splitting a skill, at least one of the child is equal or better than the skill of its parent relatively to the task. We experimentally show this in \secref{sec:perf}. The resulting curriculum can be summarized as follows: the tree will be small at the beginning, and will grow larger in the direction of feedbacks of the environment.

We now sum up the process of the \textit{tree-policy}: 1-an agent runs an episode inside the MDP of skills; the sequence of actions represents a skill; 2- the agent executes the intra-skill policy of the skill; 3- the \textit{tree-policy} is rewarded according to how well the intra-skill policy fits the task and the Q-learning applies. \ifthenelse{\equal{\conference}{icml}}{
The full algorithm of the \textit{tree-policy} is given in \appref{sec:algo}.
}

\subsection{Simultaneous training of the \textit{tree-policy} and intra-skill policies}\label{sec:catastrophic}

The MI objective requires the goal distribution to remain uniform (cf. \eqref{eq:int_reward}), however that is not our case: the agent strives to avoid some useless goals while focusing on others. In our preliminary experiments, ignoring this leads us to catastrophic forgetting of the learned skills since discriminators forget how to categorize states of the skills they never learn on. To bypass this issue and sample uniformly, we assign to each node $i$ of our tree a replay buffer containing interactions of the intra-skill policy with the environment, a RL algorithm and a discriminator ($q_{\omega}^{:i}$). At each split, intra-skill policies and buffers of a node are copied to its children; for the first node, its intra-skill policies are randomly initialized and its buffer is empty.

This way, the entire training is off-policy: the intra-skill policy fills the buffer while the discriminator and intra-skill policies learn from the interactions that are uniformly extracted from their buffers. We split the lifetime of a node into two phases: 1-the \textbf{learning phase} during which next letter's values are sampled uniformly; the \textit{tree-policy} is uniform at this node; 2-the \textbf{exploitation phase} during which the \textit{tree-policy} chooses letters with its Boltzmann policy (\secref{sec:tree_policy}).

Then, at each step, the agent runs the \textit{tree-policy} to select the discriminator in the learning phase that will learn. The discriminator samples a mini-batch of data from its children's (all leaves) buffers and learns on it. Then, all children intra-skill policies learn from the intrinsic feedback of the same interactions, output by the selected discriminator and all its parents according to \eqref{eq:myintreward}.



In addition, an hyper-parameter $\eta$ regulates the probability that each parent's discriminator learn on its children data. Their learning interactions are recursively sampled uniformly on their children. This post-exploration learning allows a node to expand its high-reward area. Without this mechanism, different uncovered states of the desired behaviour may be definitively attributed to different fuzzy goals, as shown in \secref{sec:gridworld}. 
\ifthenelse{\equal{\conference}{icml}}{
The full learning algorithm is given in \appref{sec:algolearning}.
}

Figure \ref{fig:tree} gives an example of a potential tree and how different phases coexist; the skills starting by $(0,1)$ seem to be the most interesting for the task since each letter sampling probability is high. Skills $(0,1,0)$, $(0,1,1)$, $(1,0)$ and $(1,1)$ are being learned, therefore the sampling probability of their last values is uniform.




\section{Experiments}\label{sec:expe}

The first objective of this section is to study the behavior of our ELSIM algorithm on basic gridworlds to make the visualization easier. The second purpose is to show that ELSIM can scale with high-dimensional environments. We also compare its performance with a non-hierarchical algorithm SAC \cite{haarnoja2018soft} in a single task setting. Finally we show the potential of ELSIM for transfer learning. 

\subsection{Study of ELSIM in gridworlds}\label{sec:gridworld}

\begin{figure*}[t]
\centering
\includegraphics[width=10cm]{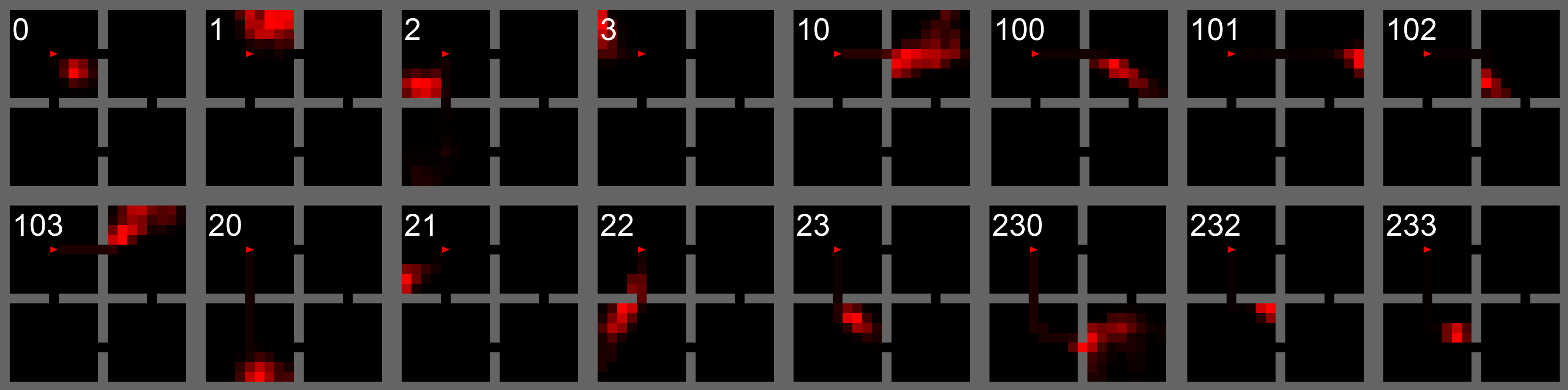}
\caption{Some skills learned by the agent in an environment composed of four rooms.}
\label{fig:rooms}
\end{figure*}

\begin{figure}[t]
\centering
\includegraphics[width=8cm]{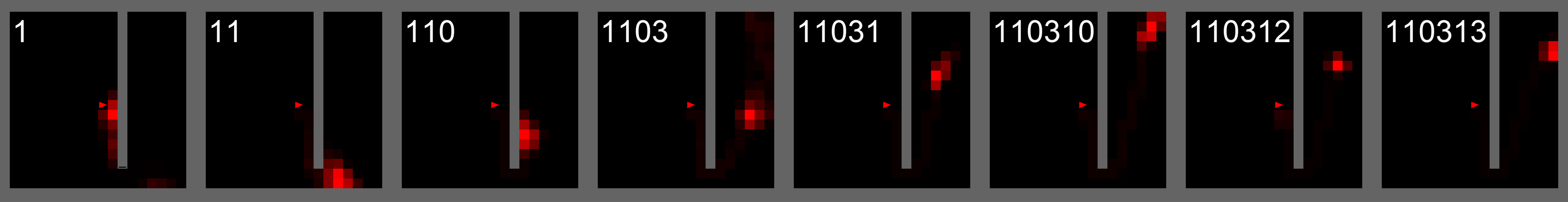}
\caption{One path in the tree of skills learned by the agent with an extrinsic reward of $1$ on the upper right side of the wall.}
\label{fig:goal_gridworld}
\end{figure}

In this section, we analyze how skills are refined on simple gridworlds adapted from gym-minigrid \cite{gym_minigrid}. Unless otherwise stated, \textbf{there is no particular task (or extrinsic reward}), thereby the \textit{tree-policy} is uniform. The observations of the agent are its coordinates; its actions are the movements into the four cardinal directions.  \ifthenelse{\equal{\conference}{icml}}{Our hyperparameters can be found in \appendixautorefname~\ref{sec:grid_hype}.} To maximize the entropy of the intra-skill policy with a discrete action space, we use the DQN algorithm \cite{mnih2015human}. The agent starts an episode at the position of the arrow (see figures) and an episode resets every 100 steps, thus the intra-skill policy lasts 100 steps. At the end of the training phase, the skills of all the nodes are evaluated through an evaluation phase lasting 500 steps for each skill. In all figures, each tile corresponds to a skill with $|V|=4$ that is displayed at the top-left of the tile. 
\figureautorefname~\ref{fig:rooms} and \ref{fig:goal_gridworld} display the density of the states visited by intra-skill policies during the evaluation phase: the more red the state, the more the agent goes over it.\\

\textit{Do the split of skills improve the exploration of an agent ?} \figureautorefname~\ref{fig:rooms} shows some skills learned in an environment of 4 rooms separated by a bottleneck.
\ifthenelse{\equal{\conference}{icml}}{The full set of skills is displayed in \appref{sec:full_4rooms}.} We first notice that the agent clearly separates its first skills $(0),(1),(2),(3)$ since the states covered by one skill are distinct from the states of the other skills. However it does not escape from its starting room when it learns these first skills. When it develops the skills close to bottlenecks, it learns to go beyond and invests new rooms. It is clear for skills $(1)$ and $(2)$ which, with one refinement, respectively explore the top-right (skill $(1,0)$) and bottom-left (skills $(2,0),(2,2),(2,3)$) rooms. With a second refinement, $(2,3,0)$ even manages to reach the farthest room (bottom-right). This stresses out that the refinement of a skill also allows to expand the states covered by the skill, and thus can improve the exploration of an agent when the rewards are sparse in an environment.\\

\begin{figure}[t]
\centering
\includegraphics[width=8cm]{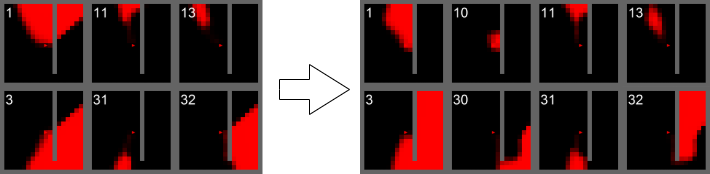}
\caption{Discriminator's probability of achieving skills $(1)$, $(3)$ and their sub-skills in every state (\textit{i.e.} $q(g|s)$). The more red the state, the more rewarding it is for the skill. The left side corresponds to the preliminary stage of the learning process (timestep $128.10^4$); the right side corresponds to the end of the learning process (timestep $640.10^4$).}
\label{fig:evolution}

\end{figure}

\textit{Do the split of skills correct a wrong over-generalization of a parent skill ?} \figureautorefname~\ref{fig:evolution} shows the evolution of the intrinsic reward function for some skills \ifthenelse{\equal{\conference}{icml}}{(see \appref{sec:full_vertical} for the full set of skills)}. The environment contains a vertical wall and settings are the same as before, except the Boltzmann parameter set to 0.5.
At the beginning, skill $(1)$ is rewarding identically left and right sides of the wall. This is due to the generalization over coordinates and to the fact that the agent has not yet visited the right side of the wall. However it is a wrong generalization because left and right sides are not close to each other (considering actions). 
After training, when the agent begins to reach the right side through the skill $(3,2)$, it corrects this wrong generalization. The reward functions better capture the distance (in actions) between two states: states on the right side of the wall are attributed to skill $(3)$ rather than $(1)$. We can note that other parts of the reward function remain identical.\\

\textit{Can the agent choose which skill to develop as a priority ?} In this part, we use the same environment as previously, but states on the right side of the wall give an extrinsic reward of $1$. Thus the agent follows the \textit{tree-policy} to maximize its rewards, using Boltzmann exploration, and focus its refinement on rewarding skills. \figureautorefname~\ref{fig:goal_gridworld} shows all the parent skills of the most refined skill which reaches $L(g)=6$. The agent learns more specialized skills in the rewarding area than when no reward is provided \ifthenelse{\equal{\conference}{icml}}{(cf. \appref{sec:expansion} for the full set of skills learned)}. \\

\textit{Summary.} We illustrated the following properties of ELSIM: 1- it expands a previously learned rewarding area when it discovers new states;  we show in \secref{sec:perf} that it improves exploration when the rewards are sparse; 2- adding letters corrects over-generalization of their parent discriminator; 3- it can focus the skill expansion towards task-interesting areas. 

\subsection{Performance on a single task}\label{sec:perf}

\begin{figure*}[t]
\centering
\includegraphics[height=3cm]{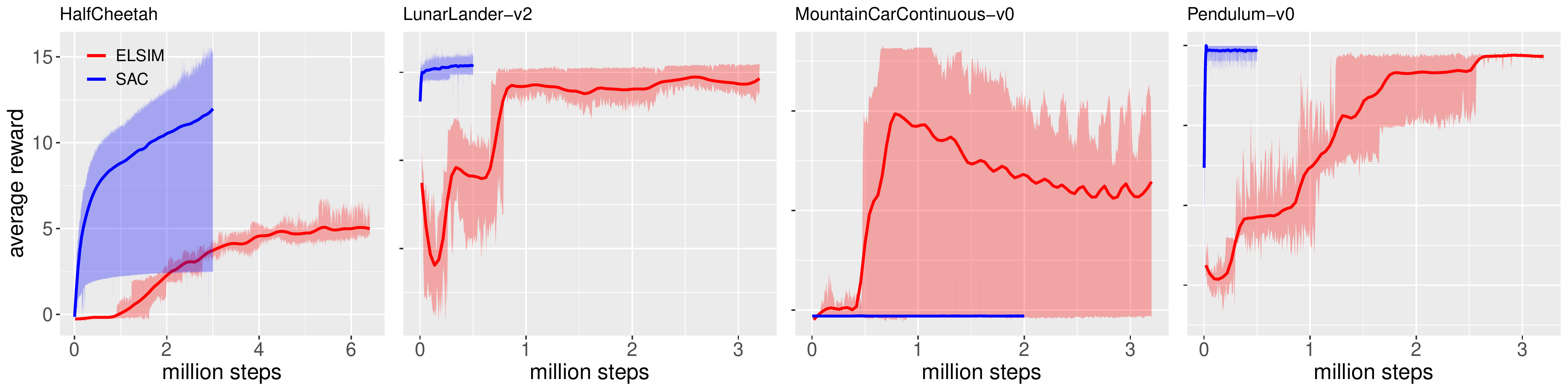}
\caption{Average reward per episode in classical environments (\textit{HalfCheetah-v2} \cite{todorov2012mujoco}, \textit{LunarLanderContinuous-v0} \cite{shariff2013lunar}, \textit{MountainCarContinuous-v0} \cite{moore1990efficient} and \textit{Pendulum-v0}) for SAC and ELSIM (averaged over 4 seeds). We use our own implementation of SAC except for \textit{HalfCheetah} for which the blue curve is the average reward of SAC on 5 seeds, taken from \cite{haarnoja2018soft}. We stopped the simulation after convergence of SAC.}
\label{fig:classics}

\end{figure*}

In this part, we study the ability of ELSIM to be competitive on high-dimensional benchmarks \footnote{\textit{HalfCheetah} has a state space and action space respectively of 17 and 6 dimensions.} \textbf{without any prior knowledge}. \ifthenelse{\equal{\conference}{icml}}{ Unless stated otherwise, used hyper-parameters can be found in \appendixautorefname~\ref{sec:mujoco_hype}.} For ELSIM, we set the maximum skill length to 10, which is reached in \textit{HalfCheetah}.

\figureautorefname~ \ref{fig:classics} respectively shows the average reward per episode for different environments. Shaded areas color are upper-bounded (resp. lower-bounded) by the maximal (resp. minimal) average reward.

First, the \textit{MountainCarContinuous} environment represents a challenge for the exploration as it is a sparse reward environment: the agent receives the reward only when it reaches the goal. In this environment, ELSIM outperforms SAC by getting a higher average reward. It confirms our results  (cf. \secref{sec:gridworld}) on \textbf{the positive impact of ELSIM on the exploration}. There is a slight decrease after reaching an optima, in fact, ELSIM keeps discovering skills after finding its optimal skill. On \textit{Pendulum} and \textit{LunarLander}, ELSIM achieves the same asymptotic average reward than SAC, even though ELSIM may require more timesteps. 
On \textit{HalfCheetah}, SAC is on average better than ELSIM. However we emphasize that \textbf{ELSIM also learns other skills}. For example in \textit{HalfCheetah}, ELSIM learns to walk and flip while SAC, that is a non-hierarchical algorithm, only learns to sprint.

\subsection{Transfer learning}\label{sec:transfer}

In this section, we evaluate the interest of ELSIM for transfer learning. We take skills learned by intra-policies in \secref{sec:perf}, reset the \textit{tree-policy} and restart the learning process on \textit{HalfCheetah} and \textit{HalfCheetah-Walk}. \textit{HalfCheetah-Walk} is a slight modification of \textit{HalfCheetah} which makes the agent target a speed of 2 \ifthenelse{\equal{\conference}{icml}}{(cf. \appref{sec:halfcheetah_walk} for more details on the new reward function)}. Intra-skill policies learning was stopped in \textit{HalfCheetah}.

\begin{figure}[t]
\centering
\includegraphics[height=3cm]{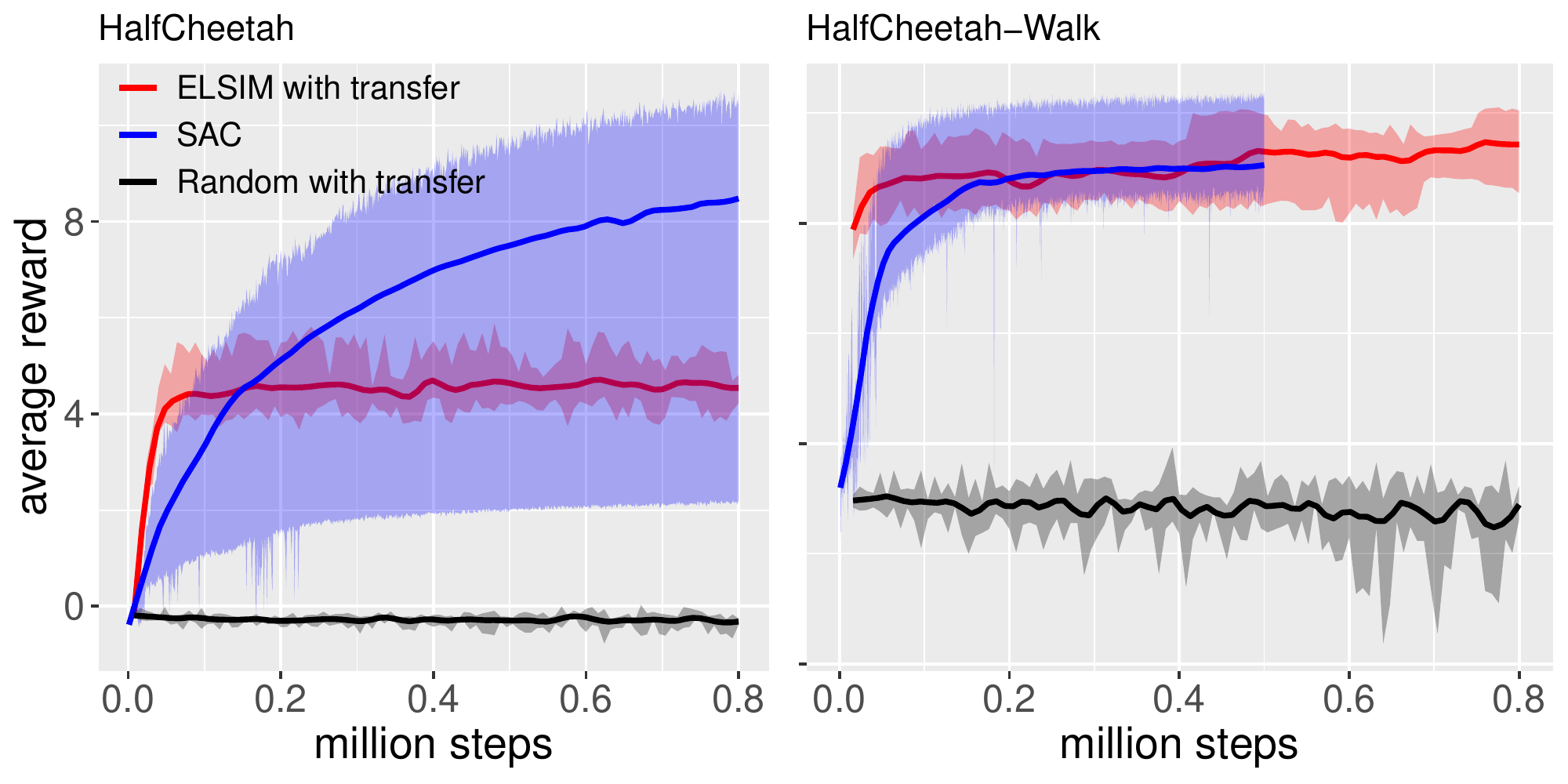}
\caption{Average reward per episode in \textit{HalfCheetah} and \textit{HalfCheetah-Walk}. We use our own implementation of SAC for \textit{HalfCheetah-Walk}. The black curve is the average reward of a random \textit{tree-policy} that uses the transfered skills.}
\label{fig:transfer}

\end{figure}

The same parameters as before are used, but we use MBIE-EB \cite{strehl2008analysis} to explore the tree \ifthenelse{\equal{\conference}{icml}}{(cf. \appref{sec:count_explore} for details)}. \figureautorefname~\ref{fig:transfer} shows that the \textit{tree-policy} learns to reuse its previously learned skills on \textit{HalfCheetah} since it almost achieves the same average reward as in \figureautorefname~\ref{fig:classics}. On \textit{HalfCheetah-Walk}, we clearly see that the agent has already learned skills to walk and that it easily retrieves them. In both environments, ELSIM learns faster than SAC, which learn from scratch. It demonstrates that skills learned by ELSIM can be used for \textbf{other tasks} than the one it has originally been trained on. 

\section{Related work}\label{sec:related}


Intrinsic motivation in RL is mostly used to improve exploration when rewards are sparse \cite{bellemare2016unifying} or to learn skills \cite{aubret2019survey}. The works that learn skills with intrinsic rewards are close to our approach and can be classified in two major categories. The first category strives to explicitly target states. The reward is defined either as a distance between agent's state and its goal state \cite{levy2018hierarchical}, or as the difference between the change in the state space and the required change \cite{nachum2019data}. However, to efficiently guide the agent, the reward functions require a good state representation \cite{nair2018visual,nachum2019near}.


\textbf{Intrinsic motivation as diversity heuristic.}
Our work mostly falls into this category, which strives to define a skill based on a MI objective (cf. \secref{sec:objmi}). Seminal works already learn a discrete set of diverse skills \cite{florensa2017stochastic,eysenbach2018diversity}. In contrast to them, we manage to learn both the skill and the skill-selection policy in an end-to-end way and we propose an efficient way to learn a large number of skills useful for the tasks the agent wants to accomplish. Recent works \cite{warde2018unsupervised,co2018self} try to learn a continuous embedding of skills, but do not integrate their work into an end-to-end hierarchical learning agent. DADS \cite{sharma2019dynamics} learn skills using a generative model over observations rather than over skills. While this is efficient on environments with simple observation space, this is computationally ineffective. In our work, rather than learning a continuous skill embedding, we strive to select and learn skills among a very large number of discretized skills. As a consequence, we focus our learned skill distribution only on task-interesting skills and we do not rely on a parametric state distribution.

\textbf{Continual learning.} Other works proposed a \textit{lifelong learning} architecture. Some assume that skills are already learned and learn to reuse them; for example H-DRLN \cite{tessler2017deep} uses a hierarchical policy to choose between ground actions and skills. They also propose to distill previously learned skills into a larger architecture, making their approach scalable. In contrast, we tackle the problem of learning skills in an end-to-end fashion, thereby our approach may be compatible. Similarly to us, CCSA \cite{kompella2017continual} addresses the catastrophic forgetting problem by freezing the learning of some experts. They mix two unsupervised learning methods to find and represent goal states, and then learn to reach them. However, their unsupervised algorithm only extracts linear features and they manually define a first set of skills. One particular aspect of continual learning is Meta-RL: how can an agent learn how to learn ? Traditional methods assume there exists a task distributions and try to generalize over it \cite{finn2017model,DuanSCBSA16}; this task distribution serves as prior knowledge. In \cite{diaynmeta}, the authors address this issue and apply MAML \cite{finn2017model} on an uniform distribution of tasks learned by DIAYN \cite{eysenbach2018diversity}. However, learning is neither focused, nor end-to-end. In the continuity of this work, CARML \cite{jabri2019unsupervised} mixes the objective of DADS \cite{sharma2019dynamics} and Meta-RL; it alternates between generating trajectories of the distribution of tasks and fitting the task distribution to new trajectories. While CARML discovers diverse behaviors with pixel-level state space, it cannot learn a global objective end-to-end like ELSIM.


\textbf{State abstraction.}
Our method can be viewed as a way to perform state abstraction \cite{li2006towards}. Rather than using this abstraction as inputs to make learning easier, we use it to target specific states. The application of our refinement method bounds the suboptimality of the representation, while the task-independent clustering ensures that skills are transferable. In contrast to our objective, existing methods usually tackle suboptimality for a task without addressing transfer learning or exploration \cite{akrour2018regularizing,DBLP:conf/icml/AbelHL16}. The \textit{k}-d tree algorithm \cite{friedman1977algorithm} has been used to perform state abstraction over a continuous state space \cite{uther1998tree}, but as above, the splitting process takes advantage of extrinsic reward and previously defined partitions are not adapted throughout the learning process. In the domain of developmental robotics, RIAC and SAGG-RIAC \cite{baranes2009r,baranes2010intrinsically} already implement a splitting algorithm building a tree of subregions in order to efficiently explore the environment and learn a forward model. More precisely, they split the state space to maximize either the sum of variance of interactions already collected or the difference of learning progress between subregions. However, these heuristics do not scale to larger continuous environments. In contrast, we assign states to subregions according to the proximity of states and use these subregions as reusable skills to solve several tasks. ASAP \cite{mankowitz2016adaptive} partitions the goal space, but does not use intrinsic motivation and the partitions are limited to hyper-plans. 




\section{Conclusion}\label{sec:conclusion}

We proposed ELSIM, a novel algorithm that continually refines discrete skills using a recently defined diversity heuristic \cite{eysenbach2018diversity}. To do so, the agent progressively builds a tree of different skills in the direction of a high-level objective. As shown in \secref{sec:expe}, \textbf{ELSIM expands the area associated to a skill} thanks to its exploratory behavior which comes from adding latent variables to the overall policy. \textbf{ELSIM also focuses its training on interesting skills relatively to some tasks}. Even though the agent is often learning a task, the skills can be defined independently from a specific task and we showed that ELSIM possibly makes them \textbf{transferable across different tasks} of a similar environment. Since the agent does not need extrinsic reward to learn, we show that it can \textbf{improve exploration} on sparse rewards environments. We believe that such a paradigm is appropriate for \textit{lifelong learning}.



Currently, our method allows to avoid the problem of catastrophic forgetting, but the counterpart is an increase of the memory footprint, which is a recurrent issue in methods based on trees. Several works addressing catastrophic forgetting may be adapted to our work, e.g. \cite{lopez2017gradient} and could potentially improve transfer learning between neural networks at different levels of our tree. In addition, ELSIM quickly gets stuck in local optimas in more difficult environments such as \textit{BipedalWalker-v2} or Pybullet environments. 
The main limitation of our approach is that we cannot select several skills in one episode, such as one would make within the \textit{option} framework \cite{sutton1999between}. To be adapted, the \textit{tree-policy} should be dependent on the true state and the diversity heuristic should maximize $\E_{s\sim \mathbb{U}(s)} I(G,S'|S)$ rather than \eqref{eq:myintreward} like in \cite{sharma2019dynamics}. Thus the curriculum algorithm should be modified. It would result that the semantic meaning of a skill would be no longer to target an area, but to produce a change in the state space. We plan to address these issues in future work.

\section{Acknowledgment}

We thank O. Sigaud and A. Dutech for their useful feedbacks on the paper. We acknowledge the support of NVIDIA Corporation with the donation of the Titan V used for this research.

\bibliographystyle{splncs}
\bibliography{references}

\appendix

\section{\textit{Tree-policy} algorithm}\label{sec:algo}

Algorithm \ref{algo:treepolicy} shows how ELSIM runs an episode in the environment without the learning part of the intra-skill policy and the discriminator. There are 3 steps: 1-an agent runs an episode inside the MDP of skills; the sequence of actions represents a skill; 2- the agent executes the intra-skill policy of the skill; 3- the \textit{tree-policy} is rewarded according to how well the intra-skill policy fits the task and the Q-learning applies.

\begin{algorithm}

\begin{algorithmic} 
\REQUIRE Environment $env$, episode length $ep_{len}$, learning rate $\phi$.
\REQUIRE Tree $T$, \textit{Tree-policy} $\pi_T$.
\REQUIRE Leaves' policies and buffers: $\forall g_{leaves} \in leaves(T),\; \pi^{g_{leaves}}_{\theta},\; \mathbb{B}^{g_{leaves}}$.

\COMMENT{It selects a skill to execute}
\STATE $node \leftarrow root(T)$
\WHILE{$not(leaf(node))$}
\STATE $node \leftarrow Boltzmann(\pi_T(node))$
\ENDWHILE
\STATE $r_{tree} \leftarrow 0$
\\
\COMMENT{It runs the intra-skill policy in the environment}
\STATE $obs \leftarrow env.reset()$
\STATE $done \leftarrow false$
\WHILE{$not(done)$} 
\STATE $action \leftarrow \pi^{node}(obs)$
\STATE $next\_obs,\,reward,\, done \leftarrow env.step(action)$
\STATE Fill $\mathbb{B}^{node}$ with $obs,done,next\_obs,action$
\STATE $obs \leftarrow next\_obs$
\STATE $r_{tree} \leftarrow r_{tree}+\gamma_T\frac{reward}{ep_{len}}$
\ENDWHILE
\\
\COMMENT{It learns Q-values of the \textit{tree-policy}}
\STATE $parent \leftarrow parent(node)$
\STATE $Q(parent,node)=(1-lr_T)\times Q(parent,node)+lr_T \times r_{tree}$
\WHILE{$not(root(node))$}
\STATE $parent \leftarrow parent(node)$
\STATE $Q(parent,node)=\max_{n \in children(node)}(Q(node,n))$
\ENDWHILE

\end{algorithmic}

\caption{\textit{Tree-policy} of ELSIM}
\label{algo:treepolicy}
\end{algorithm}

\newpage
 \section{Learning algorithm}\label{sec:algolearning}

 Algorithm \ref{algo:learning} shows one learning step in ELSIM for discriminators and intra-skill policies. The agent executes its \textit{tree-policy} on its tree; the discriminators of the intermediary encountered nodes learn with probability $\eta$ on a batch of interactions of their recursively and uniformly chosen children leaves. The last discriminator learns with probability $1$ over a batch of its leaves' interactions. This batch is labeled with the intrinsic reward computed through \eqref{eq:myintreward} and leaves' intra-skill policies learn with this batch.

\begin{algorithm}
 \begin{algorithmic} 
 \REQUIRE $batch\_size$, Tree $T$, \textit{Tree-policy} $\pi_T$.
 \REQUIRE Discriminators: $\forall g\in nodes(T),\; q_{\omega}^g$.
 \REQUIRE Leaves' policies and buffers: $\forall g_{leaves} \in leaves(T),\; \pi^{g_{leaves}}_{\theta},\; \mathbb{B}^{g_{leaves}}$.

 \COMMENT{It selects a node to learn on}
 \STATE $node \leftarrow root(T)$
 \WHILE{ $not(leaves(children(node)))$ }
 \IF{$random() < \eta $}
 \STATE $i \leftarrow 0$
 \COMMENT{Discriminators in exploitation phase learn with probability $\eta$}
 \STATE $batch \leftarrow \emptyset$
 \WHILE{$i < batch\_size$}
 \STATE $node2 \leftarrow node$ 
 \WHILE{$not(leaf(node2))$}
 \STATE $node2 \leftarrow Uniform(children(node2))$
 \ENDWHILE
 \STATE $ADD(batch,sample(\mathbb{B}^{node2}))$
 \ENDWHILE
 \STATE $Cross\_entropy(batch,q_{\omega}^{node})$
 \ENDIF
 \STATE $node \leftarrow Boltzmann(\pi_T(node))$
 \ENDWHILE

 \COMMENT{Learn the discriminator and intra-skill policies of the last node}
 \STATE $i \leftarrow 0$
 \STATE $batch \leftarrow \emptyset$
 \WHILE{$i < batch\_size$}
 \STATE $leaf \leftarrow Uniform(children(node))$
 \STATE $ADD(batch,sample(\mathbb{B}^{leaf}))$
 \ENDWHILE
 \STATE $Cross\_entropy(batch,q_{\omega}^{node})$
 \FOR{ $leaf \in children(node)$}
 \STATE Learn $\pi^{leaf}_{\theta}$ with \eqref{eq:myintreward} and SAC
 \ENDFOR
 \end{algorithmic}
\caption{Learning step of intra-skill policies and discriminators}
\label{algo:learning}
 \end{algorithm}

\section{Hyper-parameters on gridworlds}\label{sec:grid_hype}

Table \ref{hype_gridworld} shows hyperparameters used in gridworlds experiments. Hyper-parameters of the discriminator and DRL networks are identical. \textit{Tree-policy} parameters are used only when there is a high-level goal.

\begin{table}[t]
\centering
\begin{tabular}{|l|l|l|}
  \hline
    Parameters & Symbol & Values \\
  \hline
  \hline
    DRL & \\
    \hline
    Boltzmann coefficient & $\alpha_{DQN}$ & 1 \\
    Buffer size & $\mathbb{B}_{size}$ & 10k \\
    Hidden layers & $hl_{SAC}$ & 2x64 \\
    Learning rate & $lr_{DQN}$ & 0.001 \\
    Gamma & $\gamma$ & 0.98 \\
    Episode duration & $D$ & 100 \\
    Batch size & $batch\_size$ &64 \\
    Parallel environments & $n$ & 16 \\
    target smoothing coefficient &$\tau$& 0.005 \\
    \hline
    ELSIM & \\
    \hline
    Discriminators hidden layers & $lr_{D}$ & 2x64 \\
    Split threshold & $\delta$ & 0.9 \\
    Sampling probability & $\eta$  & 0.5 \\
    Average coefficient & $\beta$  & 0.02 \\
    Vocabulary size & $|V|$  & 4 \\
    Old discriminations scale & $\alpha$  & 1 \\
    \hline
    \textit{Tree-policy}& \\
    \hline
    Gamma & $\gamma_T$ & 1 \\
    Boltzmann coefficient & $\alpha_T$ & 20 \\
    Learning rate & $lr_T$ & 0.05   \\

    \hline

  \hline
\end{tabular}
 \caption{Hyper-parameters used for gridworld experiments.}
 \label{hype_gridworld}
\end{table}

\section{Hyper-parameters on continuous environments.}\label{sec:mujoco_hype}

Since the goal of ELSIM is to learn in a continual learning setting, we record the performances of ELSIM in training mode, \textit{i.e.} with its stochastic policies. Table \ref{hype_mujoco} shows hyper-parameters used in experiments on continuous environments. Learning rate of the discriminator and DRL networks are identical. In addition to these hyper-parameters, we set the weight decay of discriminators to 0.01 in the exploitation phase. We also manually scale the intra-skill rewards and entropy coefficient $\alpha_{SAC}$ to get lower-magnitude value function, thus, we divide rewards and entropy coefficient by 5 and clamp the minimal reward by (-2). SAC algorithms used by ELSIM do not use a second critic (but our implementation of SAC does). 

\begin{table}[t]
\centering
\begin{tabular}{|l|l|l|}
  \hline
    Parameters & Symbol & Values \\
  \hline
  \hline
    DRL & \\
    \hline
    Entropy coefficient & $\alpha_{SAC}$ & 0.25 \\
    Buffer size & $\mathbb{B}_{size}$ &20k \\
    SAC hidden layers & $hl_{SAC}$ & 2x128 \\
    Learning rate & $lr_{SAC}$ & 0.001 \\
    Discount & $\gamma$ &0.98 \\
    Episode duration & $D$ & 500 \\
    Batch size & $batch\_size$ &128 \\
    Parallel environments & $n$ & 16 \\
    target update parameter &$\tau$& 0.005 \\
    \hline
    ELSIM & \\
    \hline
    Discriminators hidden layers & $lr_{D}$ & 2x64 \\
    Split threshold & $\delta$ & 0.9 \\
    Sampling probability & $\eta$  & 0.5 \\
    Average coefficient & $\beta$  & 0.02 \\
    Vocabulary size & $|V|$  & 4 \\
    Old discriminations scale & $\alpha$  & 2 \\
    \hline
    \textit{Tree-policy}& \\
    \hline
    Gamma & $\gamma_T$ & 1 \\
    Boltzmann coefficient & $\alpha_T$& 5 \\
    Learning rate & $lr_T$& 0.05   \\

    \hline

  \hline
\end{tabular}

\label{hype_mujoco}
 \caption{Hyper-parameters used for experiments on continuous environments.}
\end{table}

\paragraph{Classic environments}. On MountainCar, Pendulum and LunarLander , we changed $\alpha_{sac}$ to 0.1. 

\paragraph{Our implementation of SAC}. Our implementation of SAC use the same hyper-parameters as in \cite{haarnoja2018soft}, but with a buffer size of 100000 and a learning rate of 0.001.

\paragraph{Transfer learning}. In experiments on transfer learning, we fix the number of parallel environments to 1 and reduce the length of an episode to 200. For the first 20 updates of a node, the \textit{tree-policy} remains uniform.

\section{Refining a high-stochastic skill into several low-stochastic skills}\label{sec:basic_refinement}

\begin{figure*}
\centering
\includegraphics[width=10cm]{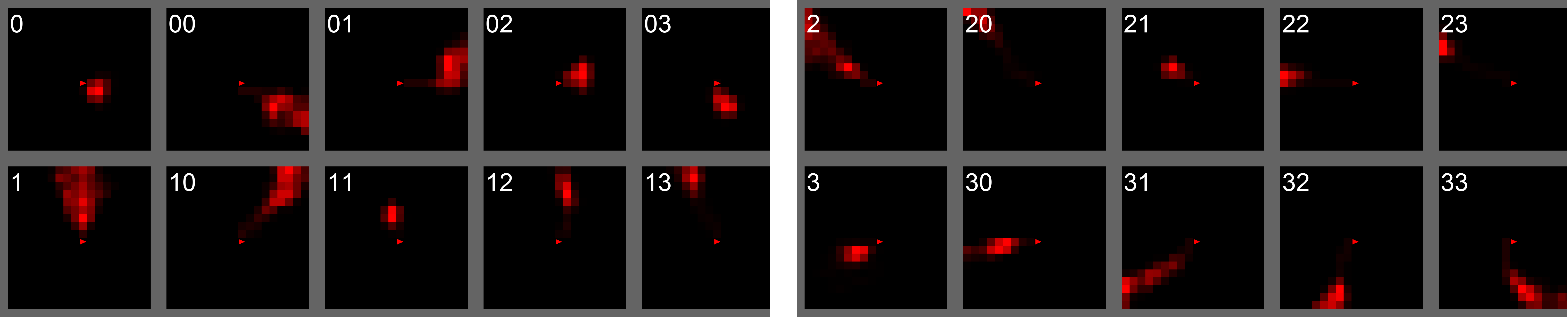}
\caption{Different states covered by the agent while doing a skill. The first and sixth columns display the intra-skill policies learned with a message length equal to 1; once the learning has been completed, the agent refines each skill into four new sub-skills, displayed on each row.}
\label{fig:refinement}

\end{figure*}

The first column of \figureautorefname~\ref{fig:refinement} shows, for each possible skill $g$ when $L(g)=1$, the states covered by the agent during the evaluation phase of the intra-skill policies. We can see that the agent clearly separates its skills since the states covered by one skill are distinct from the states of the other skill. In our example, the first skill $(0)$ makes the agent go at the right of the grid, the second one $(1)$ at the top, the third one $(2)$ at the left and the fourth one at the bottom. In contrast, columns 2-5 of Figure \ref{fig:refinement} show the intra-skill policies learned with $L(g)=2$. As evidenced by goals' numbers, the fours rightmost intra-skill policies are the refinement of the leftmost fuzzy policy on the same row. We see that the refinement allows to get lower-stochastic policies. For example, skill $(1)$ is very fuzzy while its children target very specific areas of the world. This emphasizes the benefits of using more latent variables to control the environment.

\section{Skills learned in four rooms environment}\label{sec:full_4rooms}
 
\begin{figure*}
\centering
\includegraphics[width=10cm]{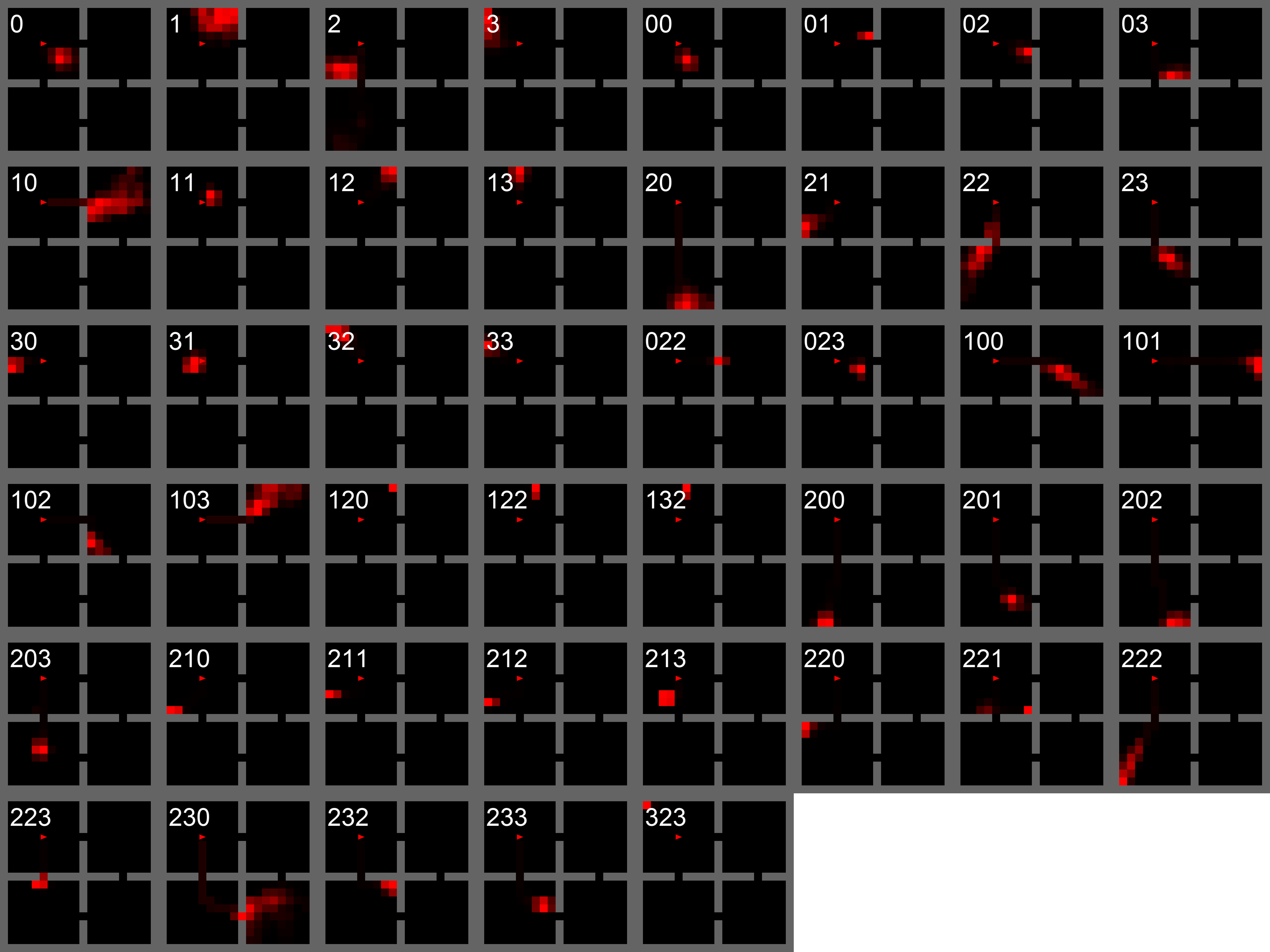}
\caption{Full set of  skills learned by the agent in an environment composed of four rooms.}
\label{fig:full_rooms}
\end{figure*}

\figureautorefname~\ref{fig:full_rooms} shows the complete set of learned skills in four rooms environment. It completes skills displayed by \figureautorefname~\ref{fig:rooms}.
 
\section{Skills learned with a vertical wall}\label{sec:full_vertical}

\begin{figure}
\centering
\includegraphics[width=8cm]{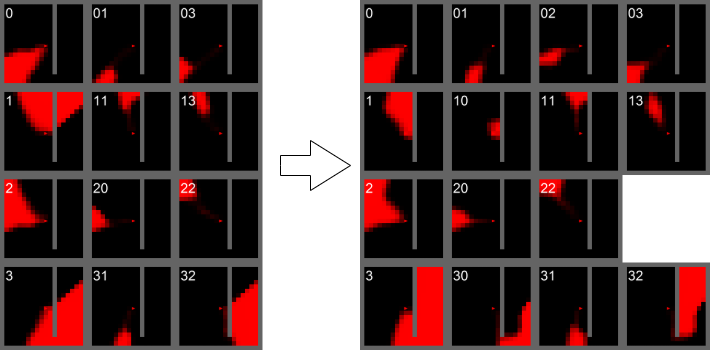}
\caption{Probability of achieving each skill in every states (\textit{i.e.} $q(g|s)$).}
\label{fig:full_evolution}
\end{figure}

\figureautorefname~\ref{fig:full_evolution} shows the evolution of the reward function for the complete set of learned skills. It completes skills displayed by \figureautorefname~\ref{fig:evolution}.

\section{Skill expansion}\label{sec:expansion}

\begin{figure}
\centering
\includegraphics[width=8cm]{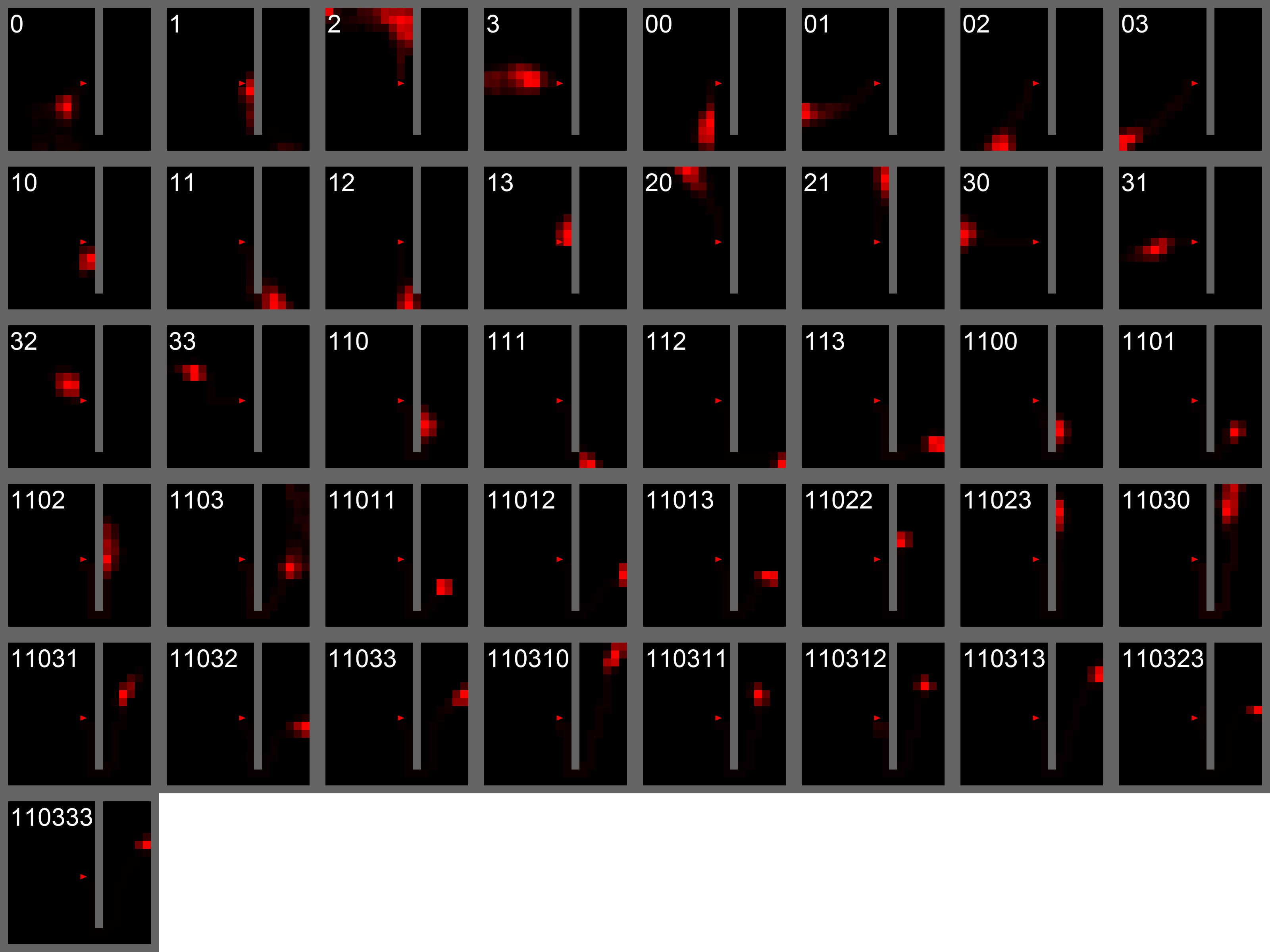}
\caption{Skills learned by the agent in an environment with a vertical wall.}
\label{fig:full_goal_gridworld}
\end{figure}

\figureautorefname~\ref{fig:full_goal_gridworld} shows the complete set of learned skills learned in an environment with a vertical wall. States on the upper right side of the wall give a reward of $1$. It completes skills displayed by \figureautorefname~\ref{fig:goal_gridworld}.

\begin{figure}
\centering
\includegraphics[width=8cm]{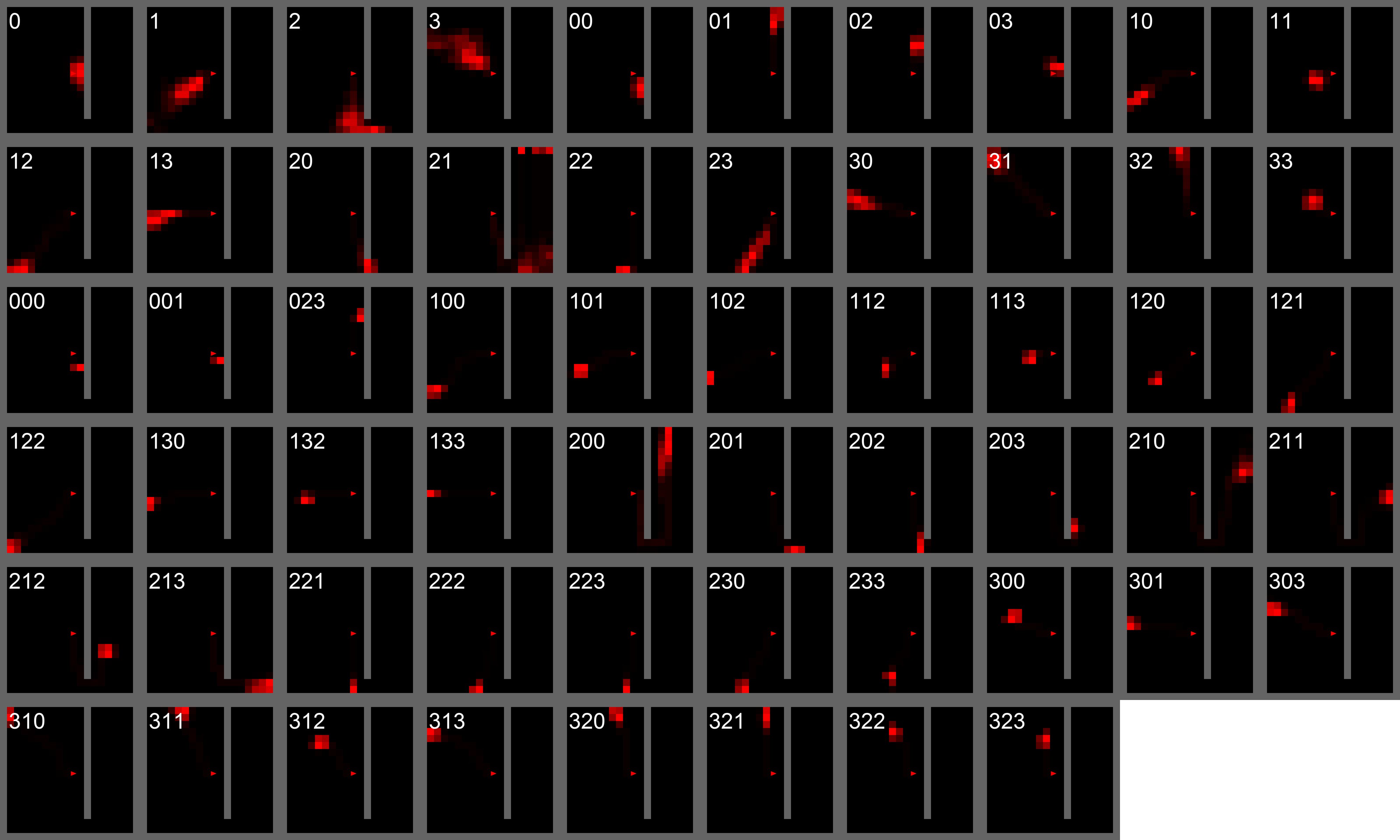}
\caption{Skills learned by the agent in an environment with a vertical wall. The agent does not focus on a specific area since there are no rewarding states.}\label{fig:nogoal_gridworld}
\end{figure}

In \figureautorefname~\ref{fig:nogoal_gridworld}, we perform the same simulation as in \figureautorefname~\ref{fig:goal_gridworld} without goals. The \textit{tree-policy} does not focus on the right side of the wall, and thus, gets less controllability than in \figureautorefname~\ref{fig:goal_gridworld}.

\section{Exploration strategy of the \textit{tree-policy} with transfer learning}\label{sec:count_explore}

To quickly learn to use skills despite their duration, we used MBIE-EB \cite{strehl2008analysis} exploration strategy. This method adds a count-based bonus to the Q-value:

\begin{equation}
    \tilde{Q}(s,a) = Q(s,a)+\frac{\beta}{\sqrt{n(s,a)}}
\end{equation}

where $n(s,a)$ is the number of time we chose $a$ in $s$ and $\beta$ is an hyper-parameter. We set it to 10 for classic HalfCheetah and 2 for walking HalfCheetah. The agent selects the action that has the larger $\tilde{Q}$.

\section{\textit{HalfCheetah-Walk}}\label{sec:halfcheetah_walk}

We introduce a slight modification of \textit{HalfCheetah-v2} to make the creature walk. The reward function used in \textit{HalfCheetah-v2}:

\begin{equation}
    R(s,a,s')=Sf(s,s')+C(a)
\end{equation}

where $C(a)$ is the cost for making moves and $Sf(s,s')$ is the speed of the agent. The reward function used in \textit{HalfCheetah-Walk} is:

\begin{equation}
    R(s,a,s')=  \left\{
      \begin{aligned}
        &Sf(s,s')+C(a)\quad &if\,Sf(s,s')>2\\
        &4-Sf(s,s')+C(a)\quad& else
      \end{aligned}
    \right.
\end{equation}

\end{document}